\pdfoutput=1
\documentclass[10pt]{article} 
\usepackage[accepted]{rlc}

\usepackage{amssymb}            
\usepackage{mathtools}          
\usepackage{mathrsfs}           
\usepackage{graphicx}           
\usepackage{subcaption}         
\usepackage[space]{grffile}     
\usepackage{url}                
\usepackage{enumitem}
\usepackage{algorithm}
\usepackage{algpseudocode}
\usepackage[nameinlink, noabbrev, capitalise]{cleveref}
\usepackage{multicol}
\crefname{assumption}{Assumption}{Assumptions}
\crefname{observation}{Observation}{Observations}

\usepackage{amsthm}

\theoremstyle{plain}
\newtheorem{theorem}{Theorem}
\newtheorem{proposition}{Proposition}
\newtheorem{lemma}{Lemma}

\newtheorem{observation}{Observation}

\theoremstyle{definition}
\newtheorem{definition}{Definition}
\newtheorem{assumption}{Assumption}
\newtheorem{example}{Example}

\theoremstyle{remark}
\newtheorem{remark}{Remark}

\DeclareMathOperator*{\argmax}{arg\,max}

\newcommand{\Real}{\mathbb{R}}
\newcommand{\CA}{\mathcal{A}}
\newcommand{\brac}[1]{\left[ #1 \right]}
\newcommand{\set}[1]{\left\{ #1 \right\}}
\DeclareMathOperator*{\E}{E}

\newcommand{\Rdag}{R^{\dagger}}
\newcommand{\belief}{\Pi_2^{b}}

\title{Inception: Efficiently Computable Misinformation Attacks on Markov Games}

\author{Jeremy McMahan  \\
    jmcmahan@wisc.edu \\
    Dept. of Computer Sciences\\
    University of Wisconsin-Madison
    \And
    Young Wu  \\
    yw@cs.wisc.edu \\
    Dept. of Computer Sciences\\
    University of Wisconsin-Madison
    \And
    Yudong Chen  \\
    yudong.chen@wisc.edu \\
    Dept. of Computer Sciences\\
    University of Wisconsin-Madison
    \And
    Xiaojin Zhu  \\
    jerryzhu@cs.wisc.edu \\
    Dept. of Computer Sciences\\
    University of Wisconsin-Madison
    \And
    Qiaomin Xie  \\
    qiaomin.xie@wisc.edu \\
    Dept. of Industrial and Systems Engineering\\
    University of Wisconsin-Madison
    }

\begin{document}

\maketitle
\allowdisplaybreaks

\begin{abstract}
    We study security threats to Markov games due to information asymmetry and misinformation. We consider an attacker player who can spread misinformation about its reward function to influence the robust victim player's behavior. Given a fixed fake reward function, we derive the victim's policy under worst-case rationality and present polynomial-time algorithms to compute the attacker's optimal worst-case policy based on linear programming and backward induction. Then, we provide an efficient inception ("planting an idea in someone's mind") attack algorithm to find the optimal fake reward function within a restricted set of reward functions with dominant strategies. Importantly, our methods exploit the universal assumption of rationality to compute attacks efficiently. Thus, our work exposes a security vulnerability arising from standard game assumptions under misinformation.
\end{abstract}

\section{Introduction}
As multi-agent systems become increasingly decentralized and privacy-focused, games with incomplete information become inevitable. In many scenarios, a player only has partial information about the opponent’s rewards and rationality, gleaned from external sources like the internet. However, misinformation spread by the opponent—possibly through fake news—can significantly impact the player’s decision-making. For example, participants in first-price auctions may intentionally misrepresent their intended bids to manipulate other bids downward. To build robust multi-agent systems, it is crucial to understand the impact of misinformation on games.

We focus on two-player Markov Games (MG). We suppose that the second player, the attacker, knows both reward functions, $(R_1, R_2)$. In contrast, the first player, the victim, only knows its reward function, $R_1$, and a misinformed attacker reward function, $\Rdag_2$. A robust victim also constructs an uncertainty set $\belief(\Rdag_2)$ of possible attacker policies. Nevertheless, the attacker can choose $\Rdag_2$ to manipulate the victim's behavior. We call these fake rewards \emph{inception attacks}. The attacker's goal is to design an inception attack that optimizes its worst-case utility.

Although inception attacks can be devastating, computing optimal attacks is often challenging. Unlike standard reward poisoning~\citep{iDSE}, an inception attack can not modify both players' rewards, which is necessary to illicit arbitrary victim behavior. Even if an oracle gave the attacker optimal fake rewards, computing a worst-case optimal attacker policy is a constrained optimization problem with nested maximins. Moreover, due to the information asymmetry, the attacker cannot utilize standard algorithms for computing robust optimization equilibrium (ROE)~\citep{RGT} or Bayes-Nash equilibrium (BNE)~\citep{BNE1} to tackle this lower-level policy optimization problem.

\paragraph{Our Contributions.} Although the computational complexity of inception might seem to limit its threat, we show that inception attacks can be efficiently computed by leveraging the universal rationality assumptions in multi-agent reinforcement learning (MARL). Specifically, for any rational or robust victim, we present an efficient algorithm for computing optimal dominant-policy inception attacks. The key insight is a rational victim always best-responds to a perceived attacker dominant strategy. Consequently, if the attacker focuses on fake reward functions admitting a dominant strategy, its complex optimization can be solved efficiently via backward induction. Our work exposes a security vulnerability arising from standard game assumptions under misinformation, motivating the need for novel approaches to building robust multi-agent systems.

To develop our inception algorithm, we first characterize outcomes in MGs with misinformation under worst-case rationality. Armed with these insights, we propose an efficient approach to compute the corresponding worst-case optimal policy for a given inception attack. Our method involves iteratively solving linear programs (LPs) based on worst-case $Q$ functions. We derive these LPs by dualizing the best-response polytope, which transforms the maximin problems into maximization problems. Our approach accommodates any finitely generated victim uncertainty set, including completely naive and secure victims.

\subsection{Related Work.}

\paragraph{Information Asymmetry.} 
Incomplete information games were first studied through the framework of Bayesian games~\citep{BNE1, BNE2, BNE3} and with the solution concept being BNE. To address the high sensitivity of BNE to the player's beliefs~\citep{BNESensExample, BNESensitivity}, the work~\citep{ExPostEquil} introduced a more robust equilibrium concept called ex-post equilibrium, which is a NE under all possible realizations of the uncertain parameters. 
Going beyond the need for belief distributions,~\citep{RGT} introduced the notion of robust games with the solution concept being ROE. However, both the BNE and ROE approaches require non-trivial assumptions about the information structure, namely, an uncertainty parametrization or distributional assumption on the opponent's rewards. Thus, they do not apply to our setting where the victim knows nothing concrete about the attacker's true rewards. 

\paragraph{Reward Poisoning Attacks.}
Most reward-poisoning attacks, for example, \citet{ma2019policy, rakhsha2020policy, rakhsha2021policy, rangiunderstanding, zhang2008value, zhang2009policy} in the single-agent setting, and \citet{wu2023reward, wu2023faking, wu2023minimally} in the multi-agent setting, focus on changing the victim's perceived rewards to induce negative behaviors rather than changing the victim's perception of the attacker's rewards. Unlike reward poisoning, which may not be possible in situations where the victim knows their preferences, inception attacks are more often possible since they fake the preferences of the attacker, which is usually not public information. Our setting also differs from past work by \citet{gleave2019adversarial, guo2021adversarial} on adversarial multi-agent reinforcement learning where an attacker is one of the agents (or controls one of the agents): they studied the problem in which an attacker modifies the action of an agent to influence the behavior of another agent (the victim). 

\subsection{Notations}
We defer formal definitions of standard concepts in game theory to \cref{sec: preliminaries}.

{\bf Normal-form Games.} Let $A \in \Real^{n \times m}$ and $B \in \Real^{n \times m}$ denote the reward matrices for the victim and attacker, respectively. We represent a pure strategy by a one-hot vector, so $e_i \in \Real^n$ corresponds to the victim's strategy $i$ and $e_j \in \Real^m$ the attacker's strategy $j$. Let $ \Delta(k) := \big\{ s \in [0,1]^k \mid \sum_{i = 1}^k s_i = 1 \big\}$ denote the set of mixed strategies, where $s \in  \Delta(k)$  corresponds to playing $e_i$ with probability $s_i$.

{\bf Markov Games.} A finite-horizon \emph{Markov game}~\citep{Markov} is defined by a tuple $G = (S,\CA,R,P,H,\mu)$ with
state-space $S$,
joint action space $\CA = \CA_1 \times \CA_2 = [n] \times [m]$ ($[i] := \set{1, \ldots, i}$),
joint reward function $R$, 
transition function $P$,
horizon $H$,
and initial state distribution $\mu$. 
We denote by $\pi = \{\pi_{1,h}(s) \in \Delta(n) \times \Delta(m)\}_{h,s}$ a joint Markovian policy. Let $\Pi_i$ denote the set of all Markovian policies for player $i\in\{1,2\}$ (victim and attacker). The value received by player $i$ under $\pi$ is the expected total rewards over $H$ steps: $V_i^{\pi} := \E^{\pi}_G\brac{\sum_{h = 1}^H \pi_{1,h}(s_h)^\top R_{i,h}(s_h)\pi_{2,h}(s_h)}$.
Similarly we define the stage value, $V_{i,h}^{\pi}(s)$, for each $h\in[H]$ by summing rewards over steps $h$ through $H$. Throughout the paper, we assume that players know the transition function $P.$

\section{Inception}\label{sec: inception}

\paragraph{Reward Uncertainty.} We formalize misinformation threats through Markov games with reward uncertainty. Suppose that the victim has learned an alleged $\Rdag_2$ directly from the attacker or external sources. A robust victim is aware that $\Rdag_2$ may be inaccurate, so it constructs an uncertainty set $\mathcal{U}(\Rdag_2)$ that it believes contains the attacker's true rewards. Furthermore, the victim believes the attacker behaves as playing some policy $\pi_2 \in \belief(\mathcal{U}(\Rdag_2))$, which depends on the belief rewards. To simplify notation, we assume the victim's belief about the attacker takes the form $\belief(\Rdag_2) \subseteq \Pi_2$, with the understanding that the victim may be using robust reasoning inside the belief function.

\begin{assumption}[Victim's Belief]\label{assum: ur}
    The victim knows some uncertain reward function $\Rdag_2$ and believes the attacker's policy must lie in the set $\belief(\Rdag_2)$. Furthermore, this is common knowledge.
\end{assumption}

\begin{example}[Naive Belief]\label{ex: naive-belief}
    If the victim believes it knows exactly which policy $\pi_2^{\dagger}$ the attacker will play, then $\belief(\Rdag_2) = \{\pi_2^{\dagger}\}$.
\end{example}

\begin{example}[Secure Belief]\label{ex: secure-belief}
    If the victim believes it knows nothing about the attacker, it may assume any attacker policy is possible, $\belief(\Rdag_2) = \Pi_2$.
\end{example}

\begin{example}[Rational Belief]\label{ex: rational-belief}
    If the victim believes the standard assumption of common-knowledge rationality, which is the case if it uses any standard MARL algorithm, then it assumes the attacker is rational. Concretely, the victim might assume the attacker plays some solution to the perceived game, $\belief(\Rdag_2) = \{\pi_2 \in \Pi_2 \mid \exists \pi_1 \in \Pi_1, (\pi_1, \pi_2) \in Sol(R_1, \Rdag_2) \}$, where $Sol$ is any standard solution concept such as DSE, NE, and maximin equilibrium\footnote{The assumption also holds for CCE, where $Sol$ corresponds to the marginal policy of the CCE for each player.}. In this work, we focus on inception attacks that only require the most basic form of rationality: rational agents never play strictly dominated strategies~\citep{iDSE}, which includes all the $Sol$ options above.
\end{example}

\subsection{Game Outcomes for Fixed $\Rdag_2$}

For any fixed $\Rdag_2$, we can reason how both players will behave when the victim believes the attacker's policy is contained in the uncertainty set $\belief(\Rdag_2)$. To formally reason about the outcomes of such games, we turn to the standard notion of worst-case rationality~\citep{RGT}. 

\begin{assumption}(Worst-Case Rationality)\label{assum: wcr}
    Both players seek to optimize their worst-case value given their available information.
\end{assumption}

\paragraph{Victim Behavior.} For the victim to be robust, it should optimize against the worst possible policy the attacker could play. By \cref{assum: ur}, it need only consider attacker policies in $\belief(\Rdag_2)$.
\begin{observation}[Victim Behaviour]\label{obs: victim}
    Under \cref{assum: ur} and \cref{assum: wcr}, the victim plays some policy $\pi_1^* \in \Pi_1^*(\Rdag_2)$ and achieves the optimal worst-case value $V_1^*(\Rdag_2)$ where,
    \begin{equation}\tag{VBR}\label{equ: wcvictim}
        \Pi_1^*(\Rdag_2) := \argmax_{\pi_1 \in \Pi_1} \min_{\pi_2 \in \belief(\Rdag_2)} V^{\pi_1,\pi_2}_1 \; \text{ and } \; V_1^*(\Rdag_2) := \max_{\pi_1 \in \Pi_1} \min_{\pi_2 \in \belief(\Rdag_2)} V^{\pi_1,\pi_2}_1.
    \end{equation}
\end{observation}
We observe that this behavior may be computationally intractable in general but is provably optimal under worst-case rationality. 
Also, this behavior can be viewed as a constrained security strategy that exploits the victim's beliefs to achieve better outcomes. This behavior directly generalizes security strategies, corresponding to the case when $\belief(\Rdag_2) = \Pi_2$. 

\paragraph{Attacker Behavior.} According to \cref{assum: ur}, the attacker knows $\belief(\Rdag_2)$. Thus, it can reason that the victim optimizes its worst-case value. Given this information, it can follow the same reasoning as the victim to predict how the victim behaves according to \cref{obs: victim}. Specifically, the attacker should choose a policy that optimizes its value for the worst possible $\pi_1 \in \Pi_1^*(\Rdag_2)$.
\begin{observation}\label{obs: attacker}
    Under \cref{assum: ur} and \cref{assum: wcr}, the attacker plays some $\pi_2^* \in \Pi_2^*(\Rdag_2)$ and achieves the optimal worst-case value $V_2^*(\Rdag_2)$ where,
    \begin{equation}\tag{ABR}\label{equ: wcattacker}
        \Pi_2^*(\Rdag_2) := \argmax_{\pi_2 \in \Pi_2} \min_{\pi_1 \in \Pi^*_1(\Rdag_2)} V^{\pi_1,\pi_2}_2 \; \text{ and } V_2^*(\Rdag_2) := \max_{\pi_2 \in \Pi_2} \min_{\pi_1 \in \Pi^*_1(\Rdag_2)} V^{\pi_1,\pi_2}_2.
    \end{equation}
\end{observation}
Importantly, the attacker exploits its information asymmetry to constrain the inner minimization. This allows the attacker to achieve a higher value than it would from a standard security strategy. 

Overall, we can see exactly how the Markov game with reward uncertainty will play out.
\begin{proposition}[Game Outcomes]\label{prop: outcomes}
    For any fixed $\Rdag_2$, under \cref{assum: ur} and \cref{assum: wcr}, $(\pi_1^*,\pi_2^*)$ is a solution to the game if and only if $(\pi_1^*,\pi_2^*) \in \Pi_1^*(\Rdag_2) \times \Pi_2^*(\Rdag_2)$.
\end{proposition}

\subsection{Inception Attacks}

The attacker can induce the fake reward $\Rdag_2$ that the victim learns, possibly by spreading misinformation. For any induced $\Rdag_2$, the attacker can achieve up to $V_2^*(\Rdag_2)$ value in the worst-case according to \cref{obs: attacker}. Thus, the attacker should choose an inception attack, $\Rdag_2$, that maximizes $V_2^*(\Rdag_2)$.

\begin{definition}[Inception]\label{def: inception}
    An \emph{optimal inception attack} is any $\Rdag_2$ that achieves $V^*_2$ where,
    \begin{equation}\tag{INC}\label{equ: inception}
        V^*_2 := \max_{\Rdag_2} V_2^*(\Rdag_2).
    \end{equation}
\end{definition}
In general, \eqref{equ: inception} is a complex, bi-level optimization problem. However, this does not mean the victim is safe from such attacks. We show in \cref{sec: algorithms} that damaging inception attacks can be computed in polynomial time for many settings. 

\begin{example}[Inception Attack]
    Consider the simple normal-form game $(R_1, R_2)$ and its corresponding inception-attack-induced game $(R_1, \Rdag_2)$ given in \cref{fig: example}. Also, suppose that the victim believes the attacker plays its part of an NE for the faked game, i.e., $\belief(\Rdag_2) = \{y \mid \exists x, \; (x,y) \in \text{NE}(R_1,\Rdag_2)\}.$
    \begin{enumerate}
        \item The original game in \cref{table: true-game} has a unique NE that is the pure strategy $(D,L)$. Thus, $\belief(R_2) = \{L\}$ and the victim plays its best-response $D$. This leads to the attacker always achieving a value of $0$.
        \item The fake game in \cref{table: fake-game} has a unique NE which is the pure strategy $(U, R)$. Thus, $\belief(\Rdag_2) = \{R\}$ and the victim plays its best-response $U$. This leads to the attacker always achieving its highest possible value of $5$ for the true game.
    \end{enumerate}
    Therefore, the attacker can simply fake that it prefers action $R$ while it actually prefers action $L$ to manipulate the victim into achieving its ideal value.
\end{example}

\begin{figure}[ht]
    \centering
    \begin{subfigure}[b]{.45\textwidth}
        \centering
        \begin{tabular}{c|c|c|}
          & L & R \\
         \hline
        U & 0, 5 & 1, 0 \\
        \hline
        D & 1, 0 & 0, 0 \\
        \hline
        \end{tabular}
        \caption{True Game}
        \label{table: true-game}
    \end{subfigure}
    \vline
    \begin{subfigure}[b]{.45\textwidth}
        \centering
        \begin{tabular}{c|c|c|}
         & L & R \\
         \hline
        U & 0, 5 & 1, 5+$\epsilon$ \\
        \hline
        D & 1, 0 & 0, $\epsilon$ \\
        \hline
        \end{tabular}
        \caption{Inception Attack}
        \label{table: fake-game}
    \end{subfigure}
    \caption{Inception Example}
    \label{fig: example}
\end{figure}

\section{Efficient Inception Algorithms}\label{sec: algorithms}

In this section, we show that for certain families of victims, the optimal inception attacks can be computed efficiently. To start, we show for a fixed $\Rdag_2$ how the attacker can efficiently compute some best response policy in $\Pi_2^*(\Rdag_2)$, which is already a complex problem. Then, we move on to computing optimal inception attacks for restricted classes of reward functions.

\subsection{Efficiently Exploiting $\Rdag_2$}\label{subsec: best-response}

Suppose that $\Rdag_2$ is fixed. We observe that computing some $\pi_2 \in \Pi_2^*(\Rdag_2)$ is a complicated optimization problem with constraints and a nested maximin optimization. Specifically,
\begin{equation}\label{equ: complex}
    \begin{split}
        \Pi_2^*(\Rdag_2) = &\argmax_{\pi_2^* \in \Pi_2} \min_{\pi_1^* \in \Pi^*_1} V_2^{\pi_1^*, \pi_2^*} \\
        &\text{s.t. } \Pi_1^* = \argmax_{\pi_1 \in \Pi_1} \min_{\pi_2 \in \belief(\Rdag_2)} V_1^{\pi_1, \pi_2}.
    \end{split}
\end{equation}
The optimization \eqref{equ: complex} can be arbitrarily complicated due to the arbitrary belief set $\belief(\Rdag_2)$. To have any hope of efficient solutions, we must restrict the belief set. Here, we consider any belief set that is a per-stage mixture of some finite set of base policies.
\begin{assumption}[Finite Generation]\label{assum: finitely-generated}
    The victim's belief set is $\belief(\Rdag_2) = \Delta(\Pi),$ where $\Pi := \{\pi_2^1, \ldots, \pi_2^K\} \subseteq \Pi_2$  is a finite set of attacker policies and $\Delta(\Pi)$ is the simplex of per-stage mixings of $\Pi$, i.e., 
    \begin{equation}
        \Delta(\Pi) := \left\{\pi \in \Pi_2 \mid \forall (h,s),\; \exists p \in \Delta(K) \; \text{ s.t.\ } \pi_{1,h}(s) = \sum_{k = 1}^K p_k  \pi_{2,h}^k(s)\right\}.
\end{equation}
    
\end{assumption}

\subsubsection{Normal-form Games}
To see how Assumption \cref{assum: finitely-generated} enables efficient computation, consider a normal-form game $(A,B)$ and $\Pi = \{y_1, \ldots, y_K\} \subseteq \Delta(m)$. 

{\bf Victim Best Response.} It is well-known \cite{NashLP} that the victim can efficiently compute a maximin solution for $A$, i.e., $\max_{x\in \Delta(n)}\min_{y\in \Delta(m)} x^{\top}A y $, by solving the LP in \cref{equ: victim-LP}. The inequalities $z\leq x^{\top} A e_j$ for all $j$ ensure that $x$ is the best response to any of the attacker's pure strategies, which then implies it is the best response to any mixture in $\Delta(m)$. In particular, $x$ must be the best response to the worst possible mixed strategy in $\Delta(m)$.

The same reasoning applies if we replace each $e_j$ with $y_j$. The inequalities $z\leq x^{\top} A y_j $ for all $j$ then guarantee that $x$ is a best response to the set $\Delta(\{y_1, \ldots, y_K\})$. Observe that we can equivalently formulate these inequalities by replacing $A$ in \cref{equ: victim-LP} with $A' := [Ay_1, \ldots, Ay_K] := A\Pi^{\top}$. Again, this implies $x$ is the best response to the worst possible mixed strategy in $\Delta(\{y_1, \ldots, y_K\})$. Since $\Pi_1^*(\Rdag_2)$ is the set of the victim's worst-case best responses to $\belief(\Rdag_2) = \Delta(\{y_1, \ldots, y_K\})$, we can compute some $x \in \Pi_1^*(\Rdag_2)$ by solving LP~\cref{equ: victim-LP} with the modified reward matrix $A'$.

\begin{figure}[ht]
    \centering
    \begin{subfigure}[b]{.48\textwidth}
        \centering
        \begin{align*}
        &\max_{x\in \mathbb{R}^{n}, z\in \mathbb{R}}\quad  \ z  \\
        \text{s.t. \quad}  &z \leq x^\top A e_j, \quad \forall j \in [m] \\
    	&      1^\top x = 1, \quad x \geq 0.
        \end{align*}
        \caption{Victim's BR LP}
        \label{equ: victim-LP}
    \end{subfigure}
    \vline
    \begin{subfigure}[b]{.48\textwidth}
        \centering
        \begin{align*}
        &\max_{y \in \Real^m, w \in \mathbb{R}^K, \alpha \in \mathbb{R}} \quad \ z^*1^\top w - \alpha\\
        \text{s.t. }\quad & \alpha + e_i^\top B y - e_i^\top A' w \geq 0 \quad \forall i \in [n]  \\
             & 1^\top y = 1, \quad y \geq 0 \quad w \geq 0.  
        \end{align*}
        \caption{Attacker's BR LP}
        \label{equ: attacker-LP}
    \end{subfigure}
    \caption{Best-response LPs}
\end{figure}

\begin{lemma}\label{lem: victim-br}
    If $(x^*,z^*)$ is a solution to LP~\ref{equ: victim-LP} for input $A' := [Ay_1, \ldots, Ay_K]$, then $V_1^*(\Rdag_2) = z^*$ and $x^* \in \Pi_1^*(\Rdag_2)$. Furthermore, $\Pi_1^*(\Rdag_2) = \{x \in \Delta(n) \mid \forall j \in [K], \; x^{\top} A' e_j \geq z^*\}$ is a non-empty polytope.
\end{lemma}

{\bf Attacker Best Response.} Now that we have understood the victim's best response $\Pi_1^*(\Rdag_2)$ polytope, the attacker can exploit this structure to compute some $y \in \Pi_2^*(\Rdag_2)$. Recall the attacker's true reward matrix is $B$. For any fixed $y$,  note that the attacker's inner minimization in \eqref{equ: complex} can be written as the  following LP and its dual in~\cref{fig: LPs2}.

\begin{figure}[ht]
    \centering
    \begin{subfigure}[b]{.45\textwidth}
        \centering
        \begin{equation*}
            \begin{split}
            \min_{x \in \Real_{\geq 0}^n} \quad & \ x^\top B y \\
            \text{s.t. } \quad & z^* - x^\top A' e_j \leq 0, \quad \forall j \in [K], \\
            	     & 1^\top x - 1 = 0.
            \end{split}
        \end{equation*}
        \caption{Primal}
        \label{equ: inter-LP}
    \end{subfigure}
    \vline
    \begin{subfigure}[b]{.53\textwidth}
        \centering
        \begin{equation*}
            \begin{split}
                \max_{w \in \Real^K_{\geq 0}, \alpha \in \Real} \quad & \ z^*1^\top w - \alpha\\
            	\text{s.t. } \quad &
            	\alpha + e_i^\top By - e_i^\top A' w \geq 0, \quad \forall i \in [n].
            \end{split}
        \end{equation*}
        \caption{Dual}
        \label{equ: dual}
    \end{subfigure}
    \caption{Attacker's Inner Minimization}
    \label{fig: LPs2}
\end{figure}

Applying $\max_{y \in \Delta(m)}$ on top of \eqref{equ: dual} yields the LP in \cref{equ: attacker-LP}, which computes a $y \in \Pi_2^*(\Rdag_2)$. We give the full derivation in the Appendix.

\begin{lemma}\label{lem: attacker-br}
    If $(y^*,w^*, \alpha^*)$ is a solution to LP~\ref{equ: attacker-LP}, then $V_2^*(\Rdag_2) = z^*1^\top w^* - \alpha^*$ and $y^* \in \Pi_2^*(\Rdag_2)$. Furthermore, $\Pi_2^*(\Rdag_2)$ is a non-empty polytope.
\end{lemma}

Therefore, the attacker can compute a $y \in \Pi_2^*(\Rdag_2)$ by first computing a solution $(x^*,z^*)$ to LP~\ref{equ: victim-LP} and then using $z^*$ to formulate and solve LP~\ref{equ: attacker-LP}. Importantly, the attacker can solve LP~\ref{equ: victim-LP} due to the information asymmetry: it knows the victim's $A$. The computation is summarized in \cref{alg: NF-BR}.

\begin{algorithm}[t]
\caption{Normal-Form Game Attacker Best Response}\label{alg: NF-BR}
    \begin{algorithmic}[1]
        \Require{$\Pi$, $A$, and $B$}
        \State $A' \gets A\Pi^{T}$
        \State $(x^*, z^*) \gets Sol(LP~\ref{equ: victim-LP}(A'))$
        \State $(y^*, w^*, \alpha^*) \gets Sol(LP~\ref{equ: attacker-LP}(z^*, A', B))$
        \State \Return $(y^*, z^*, z^*1^\top w^* - \alpha^*)$
    \end{algorithmic}
\end{algorithm}

\begin{theorem}\label{thm: nf-br}
    If $K \leq poly(m)$, then under \cref{assum: finitely-generated} the attacker can compute some $y \in \Pi_2^*(\Rdag_2)$ for a normal-form game in polynomial time by using \cref{alg: NF-BR}.
\end{theorem}

\subsubsection{Markov Games}
To extend our results to full Markov games, we solve our LPs on each stage game via backward induction. To formalize this approach, we study the worst-case stage value and its corresponding worst-case Q functions:
\begin{align}
    V^*_{1,h}(s) &:= \max_{\pi_1 \in \Pi_1} \min_{\pi_2 \in \belief(\Rdag_2)} V_{1,h}^{\pi_1, \pi_2}(s) \text{ and } V^*_{2,h}(s) := \max_{\pi_2 \in \Pi_2} \min_{\pi_1 \in \Pi_1^*(\Rdag_2)} V_{2,h}^{\pi_1, \pi_2}(s), \label{equ: V*}\\
    Q^*_{i,h}(s)[a_1,a_2] &= {R_{i,h}(s,a_1,a_2) + \sum_{s'} P_h(s' \mid s, a_1, a_2) V^*_{i,h+1}(s')}. \label{equ: Q*}
\end{align}

In particular, for each $h\in[H],s\in S$, the worst-case stage-value functions $V^*_{i,h}(s)$ can be computed from the worst-case $Q$ functions $Q^*_{i,h}(s)$,  using \cref{alg: NF-BR} with $(Q^*_{1,h}(s),Q^*_{2,h}(s))$ as the norm-form game reward matrix. We let $\pi_{1,h}(s) := \{\pi^1_{2,h}(s), \ldots, \pi^K_{2,h}(s)\}$.
\begin{lemma}\label{lem: mg-br}
    For all $h,s$, we have $(*,V^*_{1,h}(s), V^*_{2,h}(s)) = \cref{alg: NF-BR}(\pi_{1,h}(s), Q^*_{1,h}(s), Q^*_{2,h}(s))$.
\end{lemma}

Since the worst-case value is uniquely defined, we can use backward induction to compute a solution for the whole Markov game in \cref{alg: MG-BR}. 

\begin{algorithm}[t]
\caption{Markov Game Attacker Best Response}\label{alg: MG-BR}
    \begin{algorithmic}[1]
    \Require{$\Pi$ and $G$}
    \State $V^*_{i,H+1}(s) = 0$ for all $s \in S$.
     \For{$h = H$ down to $1$}
        \For{$s \in S$}
            \State $Q^*_{1,h}(s), Q^*_{2,h}(s) \gets$ \cref{equ: Q*} 
            \State $\pi_{2,h}^*(s), V^*_{1,h}(s), V^*_{2,h}(s) \gets \cref{alg: NF-BR}(\pi_{1,h}(s), Q^*_{1,h}(s), Q^*_{2,h}(s))$
        \EndFor
     \EndFor
     \State \Return $\pi_2^* := \{\pi_{2,h}^*(s)\}_{h,s}$
    \end{algorithmic}
\end{algorithm}

\begin{theorem}\label{thm: mg-br}
    If $K \leq poly(m)$, then under \cref{assum: finitely-generated} the attacker can compute some $\pi_2 \in \Pi_2^*(\Rdag_2)$ for a Markov game in polynomial time using \cref{alg: MG-BR}.
\end{theorem}

\begin{remark}[Secure Victims]
    If the victim does not trust $\Rdag_2$ as in \cref{ex: secure-belief} and simply ignores the information by computing a maximin strategy, $\max_{\pi_1 \in \Pi_1}\min_{\pi_2 \in \Pi_2} V_1^{\pi_1, \pi_2}$, the attacker can still exploit its information asymmetry. In particular, it can compute its best response in polynomial time using \cref{alg: MG-BR} on $\Pi = \{\pi_2^j\}_{j = 1}^m$ where $\pi_{2,h}^j(s) := e_j$. This leads to $\Delta(\Pi) = \Pi_2$.
\end{remark}

\subsection{Efficiently Optimizing $\Rdag_2$}\label{subsec: inception-alg}
In the previous section, we saw how to compute best-response policies for a class of beliefs of the victim. However, to compute an optimal inception attack, we require additional structure on how the victim maps rewards to belief sets. 

\begin{assumption}[Common Rationality]\label{assum: sol-belief}
     If $\pi_2^{\dagger}$ is an $\iota$-strictly dominant Markov-perfect strategy for $\Rdag_2$, then $\belief(\Rdag_2) = \{\pi_2^{\dagger}\}$. 
\end{assumption}

\begin{remark}(Rationality)
    Note that \cref{assum: sol-belief} holds whenever the victim believes common knowledge rationality as in \cref{ex: rational-belief}. We again emphasize this assumption is made by all standard MARL algorithms as rationality is the basis of these game-theoretic approaches.
\end{remark}

\paragraph{Policy Reduction.} Observe that if $\belief(R_2^{\dagger}) = \belief(R_2^{\dagger\dagger})$, then $V_2^*(R_2^{\dagger}) = V_2^*(R_2^{\dagger\dagger})$. Consequently, whenever $\belief(R_2^{\dagger}) = \{\pi_2^{\dagger}\}$, we see that $V^*_2(R_2^{\dagger})$ is completely determined by $\pi_2^{\dagger}$ and not the specific structure of $\Rdag_2$. Thus, with a slight abuse of notation, we can view $V_2^*$ as a function of $\pi_2^{\dagger}$ by defining $V^*_2(\pi_2^{\dagger}) := V^*_2(\Rdag_2)$ where $\Rdag_2$ is any reward functions satisfying $\belief(\Rdag_2) = \{\pi_2^{\dagger}\}$. Overall, we can reduce the problem of finding fake rewards to the problem of finding a fake policy. 

If $\belief(\Rdag_2) = \{\pi_2^{\dagger}\}$, then by definition $\Pi_1^*(\Rdag_2) = \argmax_{\pi_1 \in \Pi_1} V_1^{\pi_1, \pi_2^{\dagger}} =: BR(\pi_2^{\dagger})$ is just the victim's traditional best response to $\pi_2^{\dagger}$. In addition, $V_2^*(\pi_2^{\dagger}) = \max_{\pi_2 \in \Pi_2}\min_{\pi_1 \in BR(\pi_2^{\dagger})} V_2^{\pi_1, \pi_2}$ can be efficiently computed using \cref{alg: MG-BR}. 
As only deterministic policies can be dominant, this simplifies the attacker's search to a finite set. Thus, the policy version of the problem is simpler to tackle. The attacker can then do inverse reward engineering to find a reward function for which $\pi_2^{\dagger}$ is a dominant strategy, which is possible even for robust victims~\cite{iDSE}. 

\begin{lemma}[Reward-Policy Reduction] \label{lem: policy-reduction} Under \cref{assum: sol-belief},
    \begin{equation}\label{equ: policy-reduction}
        \max_{\Rdag_2 \in \mathcal{D}} V_2^*(\Rdag_2) = \max_{\pi_2^{\dagger} \in \Pi_2^D} V_2^*(\pi_2^{\dagger}),
    \end{equation}
    where $\mathcal{D}$ is the set reward functions with an $\iota$-strictly dominant Markov-perfect strategy, and $\Pi_2^D$ is the set of deterministic attacker policies. We let $\hat{V}_2:=\max_{\pi_2^{\dagger} \in \Pi_2^D} V_2^*(\pi_2^{\dagger})$ denote the optimal value. 
\end{lemma}

\cref{lem: policy-reduction} states that if the misinformation-induced reward function $\Rdag_2$ is restricted to the set admitting strictly dominant strategies, one can solve the optimal inception attack problem by solving the pure strategy optimization problem. We note this restricted set is infinite and captures many interesting reward functions.

\begin{remark}[Reward Design]
    We note the choice of $\Rdag_{2,h}(s,a) = \iota\frac{(H-h+1)(H-h+2)}{2}\mathbb{I}[a_2 = \pi_{2,h}^{\dagger}(s)]$ suffices to ensure $\pi_2^{\dagger}$ is the dominant strategy in any stage game and can be computed in polynomial time. If there are other constraints on the reward function, other reward poisoning frameworks can be used black box to compute optimal attacks.
\end{remark}

\paragraph{Algorithmic Approach.} For the normal-form game $(A,B)$, it is easy to see that for any pure strategy $j \in [m]$ that $V_2^*(j) = \max_{y \in \Delta(m)} \min_{x \in BR(j)} x^{\top} B y$ can be computed using $\cref{alg: NF-BR}(\{j\}, A, B)$ in polynomial time. The maximal pure strategy can then be found efficiently by iterating over all $j \in [m]$: $\hat{V}_2 = \max_{j} V_2^*(j)$. Thus, we can solve the policy problem for a normal-form game efficiently by repeatedly applying \cref{alg: NF-BR}.

This line of argument can be extended to Markov games by replacing $(A,B)$ with the $Q$-function matrices and using backward induction. Suppose the attacker has already constructed a partial policy $\pi_2^{\dagger}$ for times $h+1, \ldots, H$. At time $h$ and state $s$, the attacker can tentatively define $\pi_{2,h}^{\dagger}(s) = j$. For this choice, the attacker can reason about the victim's best-response set and value $\hat{V}_{1,h}(s,j)$, which is also constructed via backward induction. The attacker can then just choose the optimal $j$ that leads to its highest worst-case stage value, $\hat{V}_{2,h}(s,j)$. Formally, we define,
\begin{equation}
    \hat V_{2,h}(s) = \max_{\pi_2^{\dagger} \in \Pi_2^D} \min_{\pi_1 \in BR(\pi_2^{\dagger})} V_{2,h}^{\pi_1, \pi_2}(s)  \text{ and } \hat V_{1,h}(s) = \max_{\pi_1 \in \Pi_1} V_{1,h}^{\pi_1, \pi_2^{\dagger}}(s),
\end{equation}
to be the value of the best inception policy for the attacker at the current stage and the victim's best response value to a fixed inception policy $\pi_2^{\dagger}$, respectively. We can similarly define the corresponding $\hat{Q}$ function through \eqref{equ: Q*} by replacing $V^*$ with $\hat{V}$. Then, for any fixed $j \in [m]$, we define, 
\begin{equation}
    \hat{V}_{2,h}(s, j) = \max_{y \in \Delta(m)} \min_{x \in BR(j)} x^{\top}\hat{Q}_{2,h}(s) y \quad\text{ and } \quad\hat{V}_{1,h}(s,j) = \max_{x \in \Delta(n)} x^{\top}\hat{Q}_{1,h}(s) e_j,
\end{equation}
as the value when the attacker chooses $\pi_{2,h}^{\dagger}(s) = j$ at step $h,$ and applies the optimal inception policy for times $h+1,\ldots,H.$

\begin{lemma}\label{lem: stage-inception}
    For all $h,s,j$, we have $(*,\hat{V}_{1,h}(s,j), \hat{V}_{2,h}(s,j)) = \cref{alg: NF-BR}(\{j\}, \hat{Q}_{1,h}(s), \hat{Q}_{2,h}(s))$. Furthermore, if $j^* \in \argmax_{j \in [m]} \hat{V}_{2,h}(s, j)$, then $\hat{V}_{i,h}(s) = \hat{V}_{i,h}(s, j^*)$ for each $i \in \{1,2\}$.
\end{lemma}

In the same spirit as \cref{alg: MG-BR}, we can compute an optimal $\pi_2^{\dagger}$ using \cref{alg: inception}.

\begin{theorem}\label{thm: inception}
    Under \cref{assum: sol-belief}, \cref{alg: inception} computes a fake policy achieving value $\hat{V}_2$ in polynomial time.
\end{theorem}

\begin{algorithm}[t]
\caption{Policy Inception}\label{alg: inception}
    \begin{algorithmic}[1]
    \Require{$\Pi$ and $G$}
    \State $\hat{V}_{i,H+1}(s) = 0$ for all $s \in S$.
     \For{$h = H$ down to $1$}
        \For{$s \in S$}
            \State $\hat{Q}_{1,h}(s), \hat{Q}_{2,h}(s) \gets$ \cref{equ: Q*} 
            \For{$j \in [m]$}
                \State $\pi_{2,h}^*(s), \hat{V}_{1,h}(s,j), \hat{V}_{2,h}(s,j) \gets \cref{alg: NF-BR}(\{j\}, \hat{Q}_{1,h}(s), \hat{Q}_{2,h}(s))$
            \EndFor
            \State $\pi_{2,h}^{\dagger}(s) \gets \argmax_{j \in [m]} \hat{V}_{2,h}(s,j)$
            \State $\hat{V}_{i,h}(s) \gets \hat{V}_{i,h}(s,\pi_{2,h}^{\dagger}(s))$ for $i \in [2]$
        \EndFor
     \EndFor
     \State \Return $\pi_2^{\dagger} := \{\pi_{2,h}^{\dagger}(s)\}_{h,s}$
    \end{algorithmic}
\end{algorithm}

\begin{remark}[Dominant Mixtures]
    The algorithm can be extended to allow a mixture of a set of policies by changing $\{j\}$ to a subset of actions. This captures reward matrices with several equally dominant columns.
\end{remark}

\section{Conclusions}

In this work, we studied misinformation attacks on two-player MGs. When the victim player only knows a false attacker reward function, we showed how the game plays out under worst-case rationality. Then, we showed how the attacker can compute its worst-case optimal policy in polynomial time. Using this method as a subroutine, the attacker can exploit the universal assumption of rationality in MARL to compute an optimal dominant-policy inception attack in polynomial time. Our work highlights that the standard rationality notions produce vulnerabilities when misinformation is present. Thus, new approaches are needed to build multi-agent systems that are robust against misinformation.

\subsubsection*{Broader Impact Statement}
\label{sec:broaderImpact}
This paper presents work whose goal is to advance the field of MARL. Our work is largely theoretical, so we do not see any immediate negative societal impacts.

\subsubsection*{Acknowledgments}
\label{sec:ack}
Xie was supported in part by National Science Foundation Awards CNS-1955997 and EPCN-2339794. Zhu was supported in part by NSF grants 1836978, 2023239, 2202457, 2331669, ARO MURI W911NF2110317, and AF CoE FA9550-18-1-0166. Chen is partially supported in part by NSF grant CCF-2233152.

\bibliography{camera}
\bibliographystyle{rlc}


\appendix

\section{Extended Preliminaries}\label{sec: preliminaries}

\paragraph{Normal-form Games.} In a (finite) normal-form game, two players compete simultaneously to maximize their reward. Suppose the first player, the victim, has $n$ pure strategies and the second player, the attacker, has $m$ pure strategies. Let $A \in \Real^{n \times m}$ and $B \in \Real^{n \times m}$ denote the reward matrices for the victim and attacker, respectively. We may represent a pure strategy by a one-hot vector, so $e_i \in \Real^n$ corresponds to the victim's strategy $i$ and $e_j \in \Real^m$ the attacker's strategy $j$. Let $ \Delta(k) := \big\{ s \in [0,1]^k \mid \sum_{i = 1}^k s_i = 1 \big\}$ denote the set of mixed strategies, where choosing $s \in  \Delta(k)$  corresponds to playing $e_i$ with probability $s_i$. For a pair of mixed strategies $x \in \Delta(n)$ and $y \in \Delta(m)$, the expected rewards to the victim and attacker are $x^\top A y$ and $x^\top B y$, respectively.

\paragraph{Nash Equilibrium.} Solutions to games manifest as equilibrium concepts, among which the most famous is the \emph{Nash Equilibrium} (NE)~\citep{Nash}. An NE of a bimatrix game is a pair of strategies $(x^*,y^*) \in \Delta(n)\times\Delta(m)$ satisfying,
\begin{equation*}\label{equ: nash}
x^* \in \argmax_{x \in \Delta(n)} x^\top A y^* \;\;\text{ and }\;\; y^* \in \argmax_{y \in \Delta(m)} {x^*}^\top B y.
\end{equation*}
In words, $x^*$ and $y^*$ are mutual best-responses to each other. We let $NE(A,B)$ denote the set of all NEs for the game $(A,B)$.

\paragraph{Security Strategies.} Another solution concept is a \emph{maximin strategy} or \emph{security strategy}, which is a pair $(x^*,y^*)$ given by,
\begin{equation}
\label{equ: maximin}
x^* \in \argmax_{x \in \Delta(n)} \min_{y \in \Delta(m)} x^\top A y \;\;\text{ and }\;\; y^* \in \argmax_{y \in \Delta(m)} \min_{x \in \Delta(n)} x^\top B y.
\end{equation}
In a \emph{zero-sum} game ($B = -A$), the Minimax Theorem~\citep{maximin} implies $(x^*,y^*)$ is a NE if and only if it is a maximin strategy pair. Note that a game may have multiple NEs and maximin strategies. However, in zero-sum games, each player receives the same expected reward in every NE, which we denote by $p_v^{NE}$ and $p_e^{NE}$ respectively.

\paragraph{Markov Game Solutions.} 
Equilibrium concepts can be defined for a Markov Game by viewing it as a (very large) bimatrix game with reward matrices $(V_1^{\pi_1,\pi_2})_{\pi_1,\pi_2}$ and $  (V_2^{\pi_1,\pi_2})_{\pi_1,\pi_2}$. To avoid this complexity blowup, many works focus on \emph{Markov Perfect Equilibrum} (MPE), which requires the stricter property that a policy pair is an equilibrium at \emph{every} stage game, not just at stage $h=1$. 
Formally, $(\pi_1^*, \pi_2^*)$ is a MPE if, for all $(h,s)\in [H]\times S$, 
\begin{equation*}\label{equ: mpe}
    V_{1,h}^{\pi_1^*,\pi_2^*}(s) = \max_{\pi_1 \in \Pi_1} V^{\pi_1,\pi_2^*}_{1,h}(s)  \;\;\text{ and }\;\; V_{2,h}^{\pi_1^*,\pi_2^*}(s) = \max_{\pi_2 \in \Pi_2} V^{\pi_1^*,\pi_2}_{2,h}(s).
\end{equation*}

\section{Proofs for \texorpdfstring{\cref{sec: inception}}{Section 2}}

All the proofs from section 2 are immediate from the arguments given in the main text.

\section{Proofs for \texorpdfstring{\cref{subsec: best-response}}{Section 3.1}}

As mentioned in the main text, the proof of \cref{lem: victim-br} is immediate from standard bimatrix game theory~\citep{NashLP}.

\subsection{Proof of \texorpdfstring{\cref{lem: victim-br}}{lem: victim-br}}
The proof is immediate from the argument given in the main text.

\subsection{Proof of \texorpdfstring{\cref{lem: attacker-br}}{lem: attacker-br}}

To construct the dual in \cref{fig: LPs2}, we introduce a dual vector $w \in \Real^K_{\geq 0}$ corresponding to the inequality constraints and a dual variable $v \in \Real$ corresponding to the equality constraint. We multiply these dual variables by their respective constraints and add them to the objective to get the equivalent optimization:
\begin{equation*}
    \max_{w \geq 0,v} \min_{x \geq 0} x^\top B y + (z^* 1^\top - x^\top A')w + (x^\top 1 - 1)v
\end{equation*}
By rearranging the objective to be in terms of $x$, we get:
\begin{equation*}
    \max_{w \geq 0,v} \min_{x \geq 0} x^\top(By -A'w +1v) + z^*1^\top w - v
\end{equation*}
Moving the terms involving $x$ into the constraints then gives the Dual:
    \begin{align*}
        \max_{w\geq 0, \alpha} \quad & \ z^*1^\top w - \alpha\\
    	\text{s.t. } \quad &
    	\alpha + e_i^\top By - e_i^\top A' w \geq 0 \quad \forall i \in [n],
    \end{align*}
Applying $\max_{y \in \Delta(m)}$ outside of the Dual, yields the attacker's LP~\ref{equ: attacker-LP}:
\begin{align*}
        &\max_{y, w \in \Real^K, \alpha \in \Real} \quad \ z^*1^\top w - \alpha\\
        \text{s.t. }\quad & \alpha + e_i^\top B y - e_i^\top A' w \geq 0 \quad \forall i \in [n]  \\
        & 1^\top y = 1, \quad y \geq 0 \quad w \geq 0.  
\end{align*}

The fact that there exist optimal solutions, i.e., $\Pi_2^*(\Rdag_2) \neq \varnothing$, follows from LP~\ref{equ: attacker-LP} being feasible and bounded. Specifically, it is easily seen that choosing $y = e_1$, $w = 0$, and $\alpha = \max_{i \in [n]} |e_i^\top B e_1$ gives a feasible solution to LP~\ref{equ: attacker-LP}. Boundedness follows from the fact that by LP duality, LP~\ref{equ: attacker-LP} is value equivalent to the original problem $\max_{y \in \Delta(m)} \min_{x \in \Pi_1^*(\Rdag_2)} x^\top B y$, which is bounded being that $(A,B)$ is a finite normal-form game. This completes the proof.

\subsection{Proof of \texorpdfstring{\cref{thm: nf-br}}{thm: nf-br}}

The proof is immediate from \cref{lem: attacker-br}.

\subsection{Proof of \texorpdfstring{\cref{lem: mg-br}}{thm: mg-br}}

From \cref{thm: nf-br} and the definition of $Q^*$, it suffices to show that $V^*$ satisfies the following optimality equations:
\begin{equation}\label{equ: vic-rec}
    V^*_{1,h}(s) = \max_{\pi_{1,h}(s) \in \Delta(n)} \min_{\pi_{2,h}(s) \in \pi_{1,h}(s)} \E_{a \sim \pi_{1,h}(s)}\brac{R_{1,h}(s,a) + \sum_{s'}P_h(s' \mid s,a) V^*_{1,h+1}(s')},
\end{equation}
and,
\begin{equation}\label{equ: att-rec}
    V^*_{2,h}(s) = \max_{\pi_{2,h}(s) \in \Delta(m)} \min_{\pi_{1,h}(s) \in \Pi_{1,h}^*(s)} \E_{a \sim \pi_{1,h}(s)}\brac{R_{2,h}(s,a) + \sum_{s'}P_h(s' \mid s,a) V^*_{2,h+1}(s')},
\end{equation}
where $\Pi^*_{1,h}(s)$ is the set of maximizers to \eqref{equ: vic-rec}. This follows from similar arguments to the proof of the NashVI algorithm~\cite{NashVI} but with an added constraint set. For completeness, we give a full proof. 

\begin{proof}
    We show \eqref{equ: att-rec}. The proof of \eqref{equ: vic-rec} follows even easier as the constraint set is fixed in advance, independent of the attacker's actions.
    We proceed by induction on $h$. For the base case, consider the final time step $h = H+1$. The claim is trivial as both values are $0$.
    For the inductive step, consider any time step $h < H$ and fix any $s \in S$. Applying the bellman-consistency equations to the definition of $V^*_{2,h}(s)$ yields:
    \begin{equation*}\label{equ: cons-star}
    V^*_{2,h}(s) = \max_{\pi_2 \in \Pi_2} \min_{\pi_1 \in \Pi_1^*(\Rdag_2)}\E_{a \sim \pi_{1,h}(s)}\brac{R_{2,h}(s,a) + \sum_{s'}P_h(s' \mid s, a)V_{2,h}^{\pi}(s')}.
    \end{equation*}
    Observe that the expression decomposes: the expectation only considers the policies at the current state and time, $(\pi_{1,h}(s), \pi_{2,h}(s))$, and the summation only considers the policies at future time steps. Consequently, we can break down the $\max_{\pi_2 \in \Pi_2}$ into the separate optimizations: $\max_{\pi_{2,h}(s) \in \Delta(m)}$ and $\max_{\pi_2 \in \Pi_{2,h+1}(s')}$ for each $s' \in S$, where $\Pi_{2,h+1}(s')$ is the set of partial policies for the attacker from time $h+1$ onwards starting at state $s'$. 
    
    Similarly, we can break down the $\min_{\pi_1 \in \Pi_1^*(\Rdag_2)}$ into the separate optimizations: $\min_{\pi_{1,h}(s) \in \Pi_{1,h}^*(s)}$ and $\min_{\pi_1 \in \Pi_{1,h}^*(s')}$ for each $s' \in S$. 
    This yields the equivalent optimization:
    \begin{equation*}
        \max_{\pi_{2,h}(s) \in \Delta(m)} \max_{\pi_2 \in \bigtimes_{s'} \Pi_{2,h+1}(s')} \min_{\pi_{1,h}(s) \in \hat \Pi_{1,h}^*(s)} \min_{\pi \in \bigtimes_{s'}\Pi_{1,h}^*(s')}\E_{\pi_{1,h}(s),\pi_{2,h}(s)}\brac{\ldots}.
    \end{equation*}
    Now, consider the summation term inside of the optimization: \begin{equation*}
        \E_{\pi_{1,h}(s), \pi_{2,h}(s)} \brac{\sum_{s'} P_h(s' \mid s, a) V_{2,h+1}^{\pi}(s')}.
    \end{equation*} 
    We can apply linearity of expectation to get the equivalent term:
    \begin{equation*}
        \sum_{s'} \E_{\pi_{1,h}(s), \pi_{2,h}(s)}\brac{P_h(s' \mid s, a)V_{2,h+1}^{\pi}(s')}.
    \end{equation*}
    Also, since $V_{2,h+1}^{\pi}(s')$ depends only on the partial policies at future steps, $V_{2,h+1}^{\pi}(s')$ is constant with respect to $(\pi_{1,h}(s),\pi_{2,h}(s))$ so can be pulled out of the summation to get the equivalent term: 
    \begin{equation*}
        \sum_{s'} V_{2,h+1}^{\pi}(s')\E_{\pi_{1,h}(s), \pi_{2,h}(s)} \brac{P_h(s' \mid s, a)}.
    \end{equation*}
    Now, by the induction hypothesis, we know for any $s'$ at time $h+1$,
    \begin{align*}
        V^*_{2,h+1}(s') &= \max_{\pi_{2,h+1}(s') \in \Pi_{2,h+1}(s')} \min_{\pi_{1,h+1}(s') \in \Pi_{1,h+1}^*(s')} \\
        &\E_{\pi_{1,h+1}(s'),\pi_{2,h+1}(s')} \brac{R_{2,h+1}(s',a) + \sum_{s'} P_{h+1}(s'' \mid s', a) V^*_{2,h+2}(s'')}.
    \end{align*}
    Since the term $V_{2, h+2}^*(s'')$ is fixed and shared amongst all $s'$ at time $h+1$, we see the only variation in the stage value $V^*_{2, h+1}(s')$ comes from the choice of $(\pi_{1,h+1}(s'),\pi_{2,h+1}(s'))$ (i.e. varying the future partial policy cannot increase the objective value). These can be independently chosen for all $s'$ at time $h+1$. Thus, the optimization problems $\max_{\pi_2 \in \Pi_{2,h+1}(s')}\min_{\pi_1 \in \Pi_{1,h+1}^*(s')} V^{\pi}_{2,h+1}(s') = V^*_{2, h+1}(s')$ are separable over $s'$. Thus, we can bring the maximin over partial policies into the summation to get the term: 
    \begin{equation*}
        \sum_{s'} \max_{\nu \in \Pi_{2,h+1}(s')} \min_{\pi \in \Pi_{1,h+1}^*(s')} V_{2,h+1}^{\pi}(s')\E_{\pi_{1,h}(s), \pi_{2,h}(s)}\brac{P_h(s' \mid s, a)}.
    \end{equation*}
    Since $V^*_{2, h+1}(s') = \max_{\pi_2 \in \Pi_{2,h+1}(s')} \min_{\pi_1 \in \Pi_{1,h+1}^*(s')} V_{2,h+1}^*(s')$, the expression becomes:
    \begin{align*}
        \sum_{s'} V_{2,h+1}^*(s')\E_{\pi_{1,h}(s), \pi_{2,h}(s)}\brac{P_h(s' \mid s, a)}. 
    \end{align*}
    As $V^*_{2, h+1}(s')$ is still constant with respect to $(\pi_{1,h}(s), \pi_{2,h}(s))$, we can reverse the previous steps of pulling out this term and applying linearity of expectation to get the final expression:
    \[V^*_{2,h}(s) = \max_{\pi_{2,h}(s) \in \Delta(m)} \min_{\pi_{1,h}(s) \in \hat \Pi_{1,h}^*(s)} \E_{\pi_{1,h}(s), \pi_{2,h}(s)}\brac{R_{2,h}(s,a) + \sum_{s'} P_h(s' \mid s,a) V^*_{2, h+1}(s')}.\]
\end{proof}

\subsection{Proof of \texorpdfstring{\cref{thm: mg-br}}{thm: mg-br}}

The proof is immediate from \cref{lem: mg-br}.

\section{Proofs for \texorpdfstring{\cref{subsec: inception-alg}}{Section 3.2}}

\subsection{Proof of \texorpdfstring{\cref{lem: policy-reduction}}{lem: policy-reduction}}
The proof is immediate from the argument given in the main text.

\subsection{Proof of \texorpdfstring{\cref{lem: stage-inception}}{lem: stage-inception}}

The proof follows similarly to the proof of \cref{lem: mg-br} and the arguments from the main text.

\subsection{Proof of \texorpdfstring{\cref{thm: inception}}{thm: inception}}

The proof is immediate from \cref{lem: stage-inception}.

\end{document}